%% file: main.tex
\documentclass[letterpaper, 10 pt, conference]{ieeeconf}  

\IEEEoverridecommandlockouts       

\usepackage[backend=biber,
            url=false,
            isbn=false,
            doi=false,
            backref=false,
            style=ieee,
            natbib=true,
            mincitenames=1,
            maxcitenames=1,
            citestyle=numeric-comp,
            sorting=nyt,
            block=none]{biblatex}

\renewcommand{\bibfont}{\small}

\input{0-preamble.tex}
\input{0-defs.tex}

\title{\LARGE \bf
All You Need is LUV: Unsupervised Collection\\ of Labeled Images Using UV-Fluorescent Markings}

\author{
Brijen Thananjeyan\beetle{}, Justin Kerr\beetle{}, 
Huang Huang, Joseph E. Gonzalez, Ken Goldberg
\thanks{\beetle{} Equal contribution}
\thanks{The AUTOLab at UC Berkeley (\href{mailto:automation@berkeley.edu}{automation@berkeley.edu})}}

\begin{document}

\maketitle

\begin{abstract}
Large-scale semantic image annotation is a significant challenge for learning-based perception systems in robotics. Current approaches often rely on human labelers, which can be expensive, or simulation data, which can visually or physically differ from real data. This paper proposes \algname{}, a novel framework that enables rapid, labeled data collection in real manipulation environments without human labeling. \algabbr{} uses transparent, ultraviolet-fluorescent paint with programmable ultraviolet LEDs to collect paired images of a scene in standard lighting and UV lighting to autonomously extract segmentation masks and keypoints via color segmentation. We apply \algabbr{} to a suite of diverse robot perception tasks to evaluate its labeling quality, flexibility, and data collection rate. Results suggest that \algabbr{} is 180-2500 times faster than a human labeler across the tasks. We show that \algabbr{} provides labels consistent with human annotations on unpainted test images. The networks trained on these labels are used to smooth and fold crumpled towels with 83\% success rate and achieve 1.7mm position error with respect to human labels on a surgical needle pose estimation task. The low cost of \algabbr{} makes it ideal as a lightweight replacement for human labeling systems, with the one-time setup costs at \$300 equivalent to the cost of collecting around 200 semantic segmentation labels on Amazon Mechanical Turk. Code, datasets, visualizations, and supplementary material can be found at \url{https://sites.google.com/berkeley.edu/luv}.
\end{abstract}

\input{1-introduction}

\input{2-related-work}

\input{3-alg}
\input{4-experiments}
\input{5-discussion}

\input{6-acknowledgement}

\renewcommand*{\bibfont}{\footnotesize}
\printbibliography
\clearpage

\end{document}

%% file: 0-preamble.tex
\usepackage{graphics}
\usepackage[pdftex]{graphicx}
\usepackage{wrapfig}
\DeclareGraphicsExtensions{.pdf,.png,.jpg}
\usepackage{epsfig}
\usepackage[font={small}]{caption}
\usepackage{subcaption}
\usepackage[rightcaption]{sidecap}
\usepackage{pbox}

\usepackage{bigstrut}
\usepackage{makecell}

\usepackage{mathtools}
\usepackage{amssymb,amsmath}
\usepackage{gensymb} 
\usepackage{nicefrac}       
\numberwithin{equation}{section} 
\usepackage{algorithm}
\usepackage{algpseudocode}

\usepackage{textcomp} 

\usepackage{array} 
\usepackage{tabularx}
\usepackage{multirow}
\usepackage{multicol}
\usepackage{booktabs}
\usepackage{tabulary}

\usepackage[utf8]{inputenc}
\usepackage{units}
\usepackage{bm}
\usepackage{xspace}
\usepackage{flushend}
\usepackage{balance} 
\usepackage{csquotes}
\usepackage{makeidx}
\usepackage{blindtext}

\usepackage{url}

\usepackage{hyperref}
\hypersetup{
    colorlinks=true,
    linkcolor=black,
    citecolor=black,
    filecolor=cyan,
    urlcolor=black
}

\renewcommand{\baselinestretch}{1.0}

%% file: 0-defs.tex
\newcommand{\algabbr}{LUV}
\newcommand{\algname}{Labels from UltraViolet (LUV)}

\DeclareRobustCommand{\beetle}{%
  \begingroup\normalfont
  \includegraphics[height=\fontcharht\font`\B]{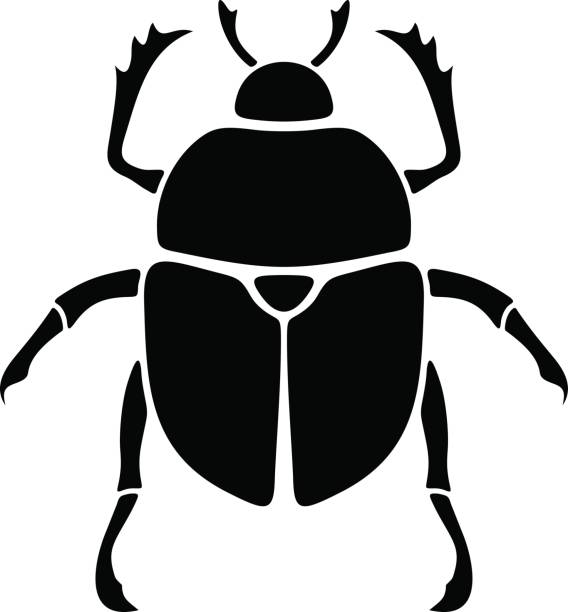}%
  \endgroup
}

%% file: 1-introduction.tex
\section{Introduction}
Supervised learning is a popular technique for training perception and planning systems for robots, with encouraging results in applications such as autonomous driving~\cite{geiger2013vision,kollar2021simnet,scale_ai}, robot object grasping~\cite{kollar2021simnet,danielczuk2019segmenting,florence2018dense,mahler2019learning,huang2022mechanical}, deformable manipulation~\cite{sundaresan2020learning,ganapathi2021learning,seita_fabrics_2020,yan2020self,zhang2021robots,grannen2020untangling,lim2021planar}, and robot-assisted surgery~\cite{paradis2020intermittent,hwang2020efficiently,wilcox2021learning}.
Supervised learning requires labeled data, and a common approach is for humans to hand-label images with segmentation masks, keypoints, and class labels~\cite{wilcox2021learning,grannen2020untangling,ji2018learning}. 
However this is time-consuming, error-prone, and expensive~\cite{scale_ai}, especially when dense or 3-D annotations are required~\cite{sundaresan2020learning,ganapathi2021learning,kollar2021simnet,danielczuk2019segmenting,florence2018dense}.
An alternative approach is to use simulated data, where data annotation can be densely and autonomously generated at scale at relatively low cost~\cite{huang2021mechanical,huang2022mechanical,mahler2019learning,kollar2021simnet,sundaresan2020learning,danielczuk2019segmenting,ganapathi2021learning}.
However, sim-to-real transfer is still an active area of research with many current limitations and open questions~\cite{zhao2020sim}. Many applications would benefit from an efficient method to collect and annotate real image annotations without human supervision.

\begin{figure}[h!]
    \centering
    \includegraphics[width=\columnwidth]{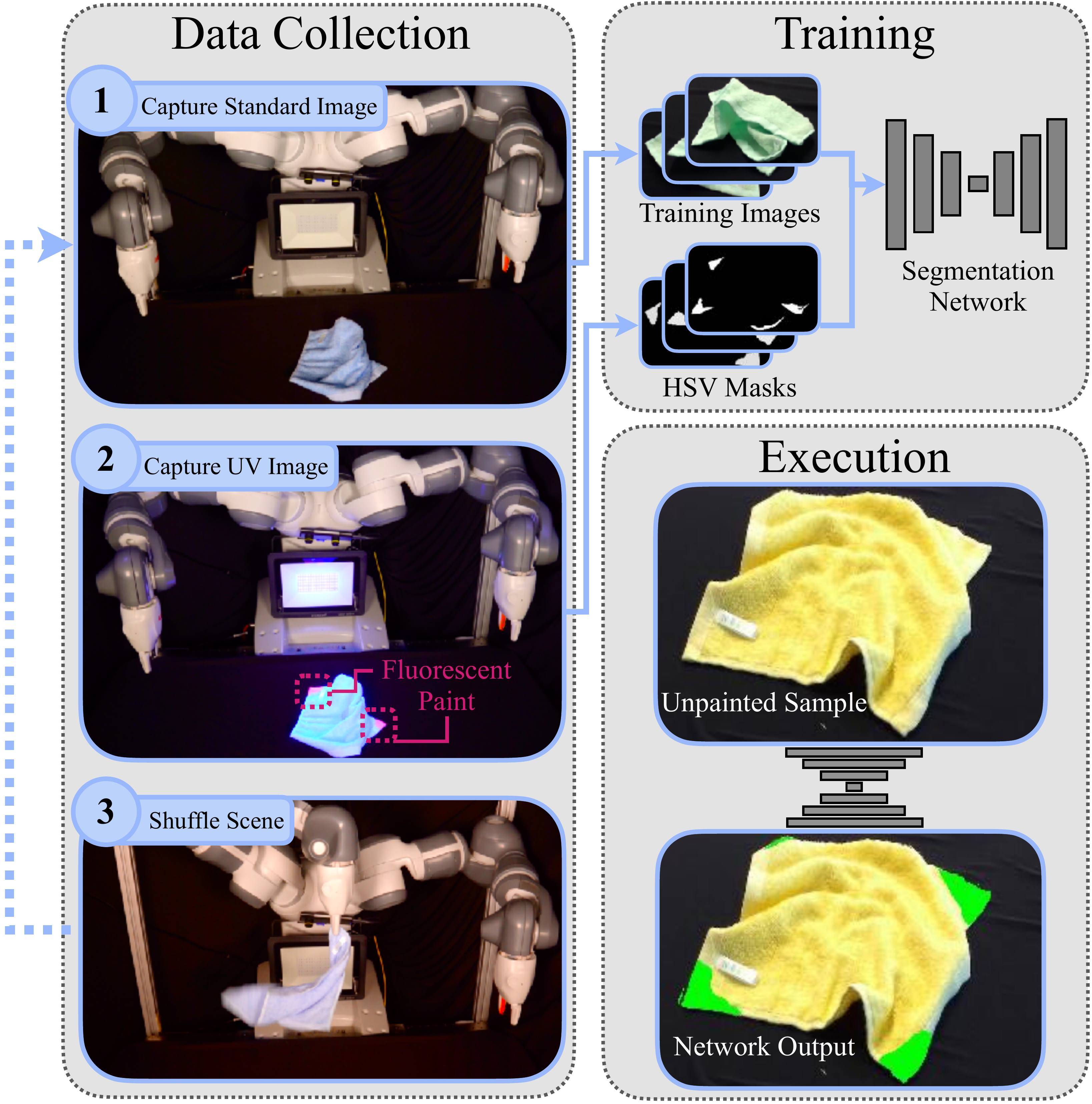}
    \caption{\textbf{Framework overview.} \textbf{Data collection:} \algabbr{} collects paired images in standard and UV lighting. Relevant keypoints and objects are coated with transparent, UV-fluorescent markings which are used to extract annotations from the UV images for the standard images. \textbf{Training:} The annotations are used to train a segmentation network to predict  masks from images under standard lighting. In contrast to prior approaches to obtain segmentation labels for images, \algabbr{} requires no human annotator or simulator. \textbf{Execution:} During execution, the trained network is evaluated on standard images of unpainted objects.}
    \label{fig:algo}
\end{figure}

In this paper, we present \algname{} (Figure~\ref{fig:algo}), a novel framework for rapidly collecting ground-truth annotations without human labels. \algabbr{} uses an array of ultraviolet lights placed around a manipulation workspace that can be switched automatically. We mark objects or keypoints in the scene with transparent, ultraviolet fluorescent paints that are nearly invisible in visible light but highly visible under ultraviolet radiation. For physical configurations, \algabbr{
} takes two images: one with standard lighting and one with the ultraviolet lights turned on. \algabbr{} provides precise labels for the standard image by using the paired ultraviolet image and trains a network on the resulting dataset to make predictions on subsequent scenes under standard lighting. 
\algabbr{} has several desirable qualities of a scalable system for ground-truth data collection:
\begin{enumerate}
    \item \textbf{accurate} segmentation masks and keypoints on \textit{real images} that do not alter the physical appearance,
    \item \textbf{flexibility} to a wide variety of materials and tasks,
    \item \textbf{rapid} data collection with \textit{no human annotation,}
    \item \textbf{inexpensive} setup with off-the-shelf parts costing less than \$300 total.
    \item \textbf{unambiguous} labels with no human subjectivity~\cite{cordts2016cityscapes}.
\end{enumerate}

To quantitatively measure these properties, we apply \algabbr{} to 3 real-world perception tasks in robot manipulation: locating fabric keypoints~\cite{seita_fabrics_2020,kollar2021simnet,ganapathi2021learning,seita2018deep}, cable segmentation~\cite{sundaresan2020learning}, and surgical needle segmentation~\cite{wilcox2021learning,sen2016automating,sundaresan2019automated}. For each, we report the speed of data collection, qualitative invisibility of markings, transferability to unpainted images, and correspondence to human ground truth labels. Because we are able to collect data much more efficiently than in prior work, we study more visually complex variants of these problems than previously considered. This paper is not meant to improve on state-of-the art performance on these tasks, but rather to provide a new labeling method that can be applied to a variety of problem domains. 

This paper makes the following contributions:
\begin{enumerate}
    \item \algabbr{}, an easy-to-setup, inexpensive framework for quickly collecting accurate ground truth image annotations. We provide detailed instructions on how to replicate this system and best practices for applying it to different scenarios.
    \item A user-friendly codebase for running \algabbr{} and training segmentation networks.
    \item Experimental results evaluating its flexibility and performance on 3 real-world robot perception tasks, suggesting that networks trained with data from \algabbr{} can make reliable predictions on unpainted objects. We report intersection over union (IOU) metrics showing \algabbr{} produces labels close to humans', enables 83\% task success on folding towels from corner detections, and localizes needles within 1.7mm of human labels.
    \item Publicly-available, annotated datasets for fabric corner keypoints (3640 labeled images), cable segmentation masks (486 labeled images), and needle segmentation masks (1364 labeled images).
\end{enumerate}

%% file: 2-related-work.tex
\section{Related Work}\label{sec:related_work}

\subsection{Self-supervised Robot Data Collection and Labeling}
To alleviate the need for explicit human labels, many prior works have discovered creative methods to automate data collection and labeling, often leveraging the structure of the task. This is commonly applied when training dynamics models that predict the resulting state after an action, both on images~\cite{hoque2020visuospatial,finn2017deep,xie2019improvisation,dasari2019robonet} and lower-dimensional state such as keypoints~\cite{hwang2020efficiently,manuelli2020keypoints}. Recently, self-supervision has been applied in imitation learning to obtain ground-truth action labels for image-based policies by manually resetting the robot to a goal or known configuration, perturbing the end effector by a known displacement, and using the displacement with the initial pose to compute an action label~\cite{wilcox2021learning,di2022learning,ma2021extrinsic}. Self-supervision is also a popular technique in reinforcement learning when automatic resets are readily available. \citet{kalashnikov2018qt} and \citet{pinto2016supersizing} use the result of autonomously explored robot grasps to supervise a grasp quality estimator, making it possible for them to collect 580K and 50K physical grasps respectively. \algabbr{} makes state/label estimation possible in situations where autonomous labels were previously difficult to generate and can be applied to extend the above self-supervision techniques. \algabbr{} is most similar to \citet{qian2020cloth}, who use visible markers to label images for a network that predicts cloth features from depth images alone. In constrast to this work, \algabbr{} can be used to label RGB images, which is useful in tasks such as needle segmentation where active depth sensors tend to fail~\cite{wilcox2021learning}.


\subsection{Fluorescent Marking Technology}
Fluorescent markings are commonly used in non-robotics applications to track and identify target objects. In medicine, near-infrared (NIR) fluorescence is used for mapping, tumour identification, and angiography in cancer surgery~\cite{schaafsma2011clinical}. In water treatment applications, fluorescence spectroscopy is applied to identify fouling agents and monitor wastewater quality quality~\cite{carstea2016fluorescence}. UV-fluorescent paints have been used for tracking cloth state during robot manipulation~\cite{siyu2012evaluation}. In contrast, we do not use UV-fluorescence at execution time, and instead we use the fluorescent markings to label training data for cloth tracking. 

\subsection{Semantic Segmentation}
Semantic segmentation is a well-studied field in computer vision with significant advances in the past decade due to the emergence of large labeled datasets like COCO, PASCAL VOC, CityScapes~\cite{lin2014microsoft,cordts2016cityscapes,everingham2010pascal}, and the development of segmentation architectures like fully-convolutional networks (FCNs)~\cite{long2015fully}, U-Nets~\cite{ronneberger2015u}, and region-proposal networks~\cite{ren2015faster}. Training these networks relies on a large dataset of images with pixel-wise annotations of objects. Previous work accomplishes this with cloud-based image labeling, which distributes the task of labeling data to human laborers on platforms like Amazon Mechanical Turk or Scale~\cite{scale_ai,turk}. This method, though effective, suffers from inconsistent label quality, difficulty in specifying labels in ambiguous situations, and cumbersome oversight processes to filter low quality labels. In addition, for involved tasks like image segmentation, the recommended price of a label on Turk is \$0.82~\cite{turk} is expensive for large datasets.

%% file: 3-alg.tex
\section{\algname{}}\label{sec:alg}
In this section, we present \algname{}, a framework for generating image annotations in manipulation domains without human labeling.

\subsection{Framework Overview}\label{subsec:framework}
\algabbr{} is comprised of two phases: \textbf{training} and \textbf{execution}.

\subsubsection{Training}
During training, relevant keypoints and segmentation masks are coated with transparent UV fluorescent paint, that are nearly invisible in natural lighting but brightly light up in different colors under UV radiation. The robot collects a dataset of paired images in the workspace: $\mathcal{D}_{\rm train} = \left\{\left(I_{i, \rm std}, I_{i, \rm uv}\right)_{i=1}^{N_{\rm train}}\right\}$. Each standard RGB image $I_{i, \rm std}\in \mathbb{R}^{H\times W \times 3}$ is taken under standard workspace lighting conditions. For the UV RGB images, $I_{i, \rm uv}\in \mathbb{R}^{H\times W \times 3}$, the workspace is illuminated by ultraviolet spectrum LED lamps. The workspace is otherwise unmodified between the paired images. The fluorescence is used to extract segmentation masks and keypoints from the UV images, which are used as training labels for the standard images. A learning-based perception model $f_{\theta}$ is trained on the labeled dataset.

\subsubsection{Execution}
During execution, the perception model $f_\theta$ is evaluated only on images under standard workspace lighting conditions containing unpainted objects.

\subsection{UV Fluorescent Paint}\label{subsec:paint}
\begin{figure}[t]
    \centering
    \includegraphics[width=\columnwidth]{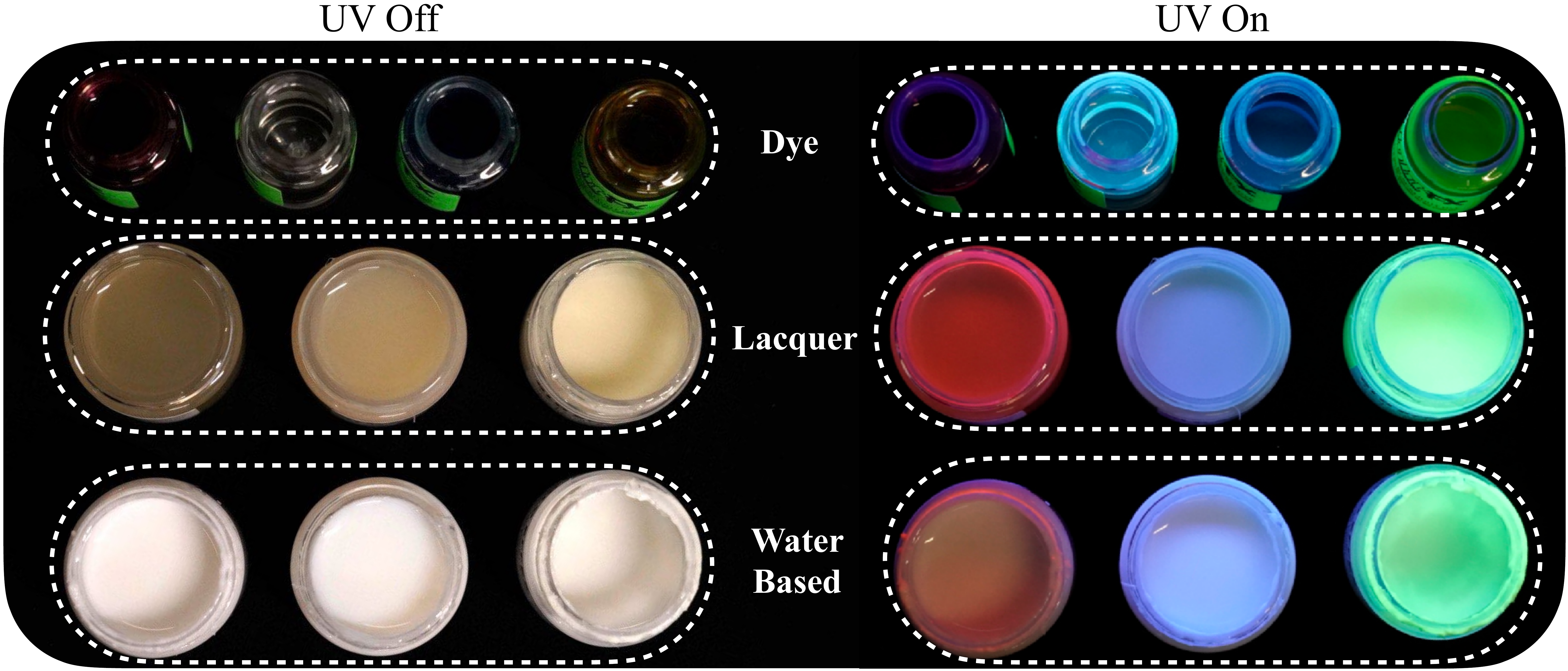}
    \caption{\textbf{UV Paint Types}: We consider three types of UV-fluorescent paint in this paper: dyes (top), lacquer-based paint (middle), and water-based paints (bottom). We describe their properties and painting techniques in Section~\ref{subsec:paint}. Most paints dry almost clear.}
    \label{fig:paints}
\end{figure}

We experiment with three types of fluorescent marking substances (Figure~\ref{fig:paints}) and describe the most successful techniques for applying them, as well as their robustness and surface finish properties.

\begin{figure}[t]
    \centering
    \includegraphics[width=\columnwidth]{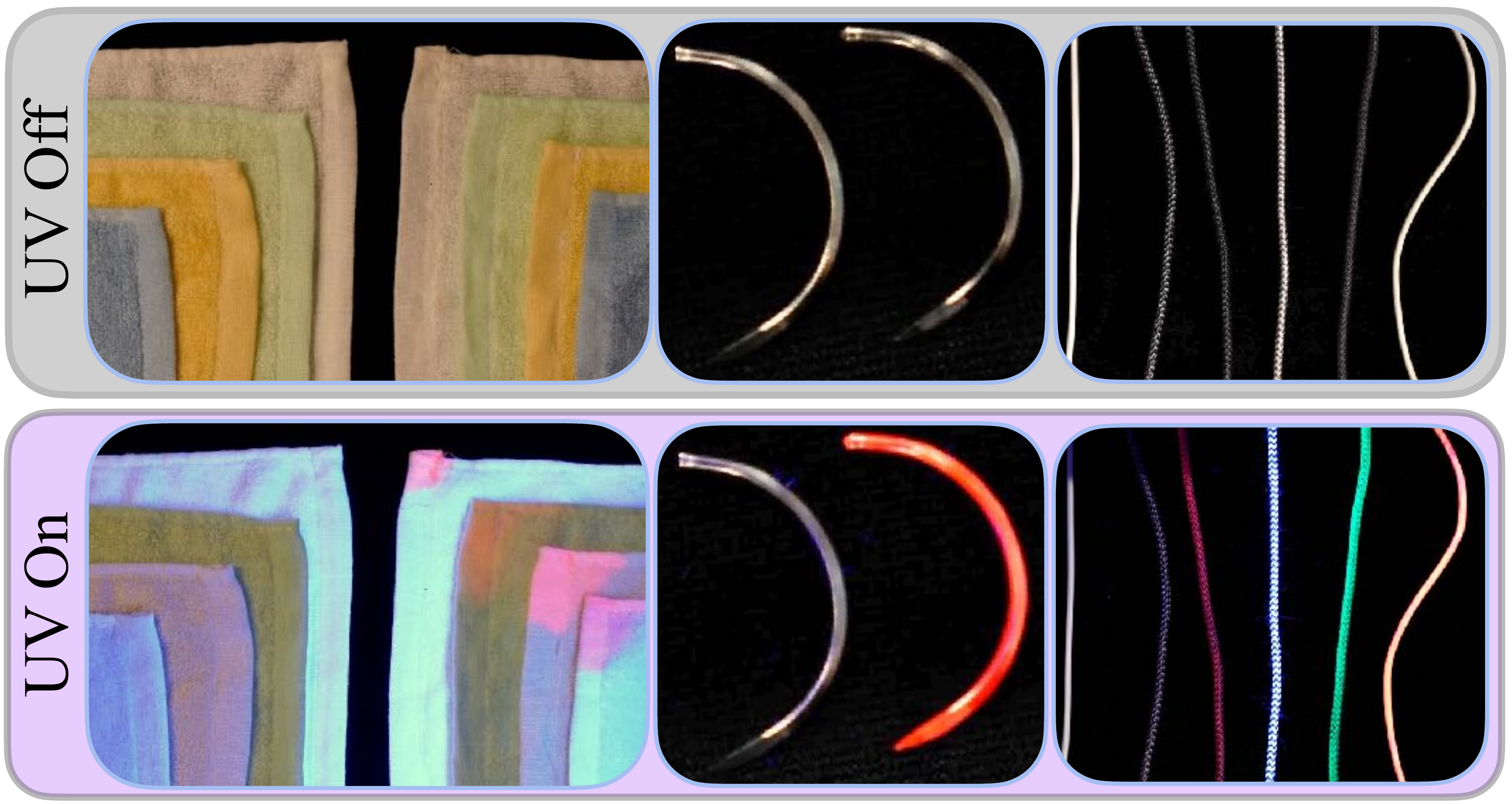}
    \caption{\textbf{UV Paint Transparency}: The top row of this figure contains both painted and unpainted towels, surgical needles, and household charging cables under standard lighting. The left objects in each image are unpainted and the right objects are painted with transparent UV paint. Under standard lighting, the painted objects are difficult to visually distinguish from the unpainted objects, without careful inspection. Under UV radiation (bottom row), the painted objects distinctly fluoresce based on the color of the paint used.}
    \label{fig:paint}
\end{figure}

\begin{figure*}[t]
    \centering
    \includegraphics[width=\textwidth]{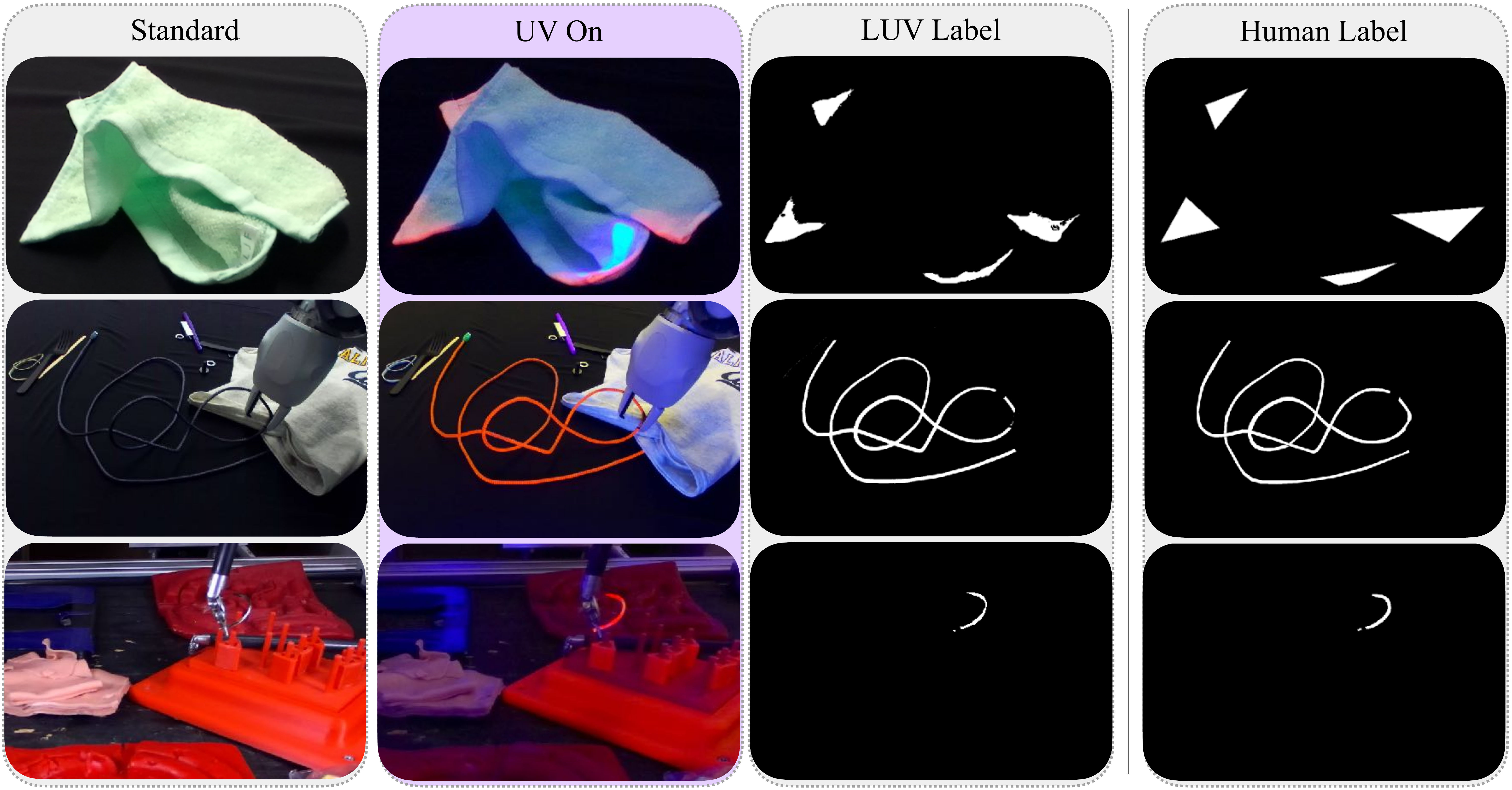}
    \caption{\textbf{LUV Data Collection}: We consider three tasks from prior robot manipulation literature: cloth corner detection, cable segmentation, and needle segmentation. To collect a labeled datapoint, \algabbr{} collects an image in standard lighting and then collects an image of the same scene under UV lighting. Then, color segmentation is used to extract the relevant annotations for the image. HSV filtering described in Section~\ref{subsec:mask_generation} is able to extract the red fluorescence of the needle without capturing any of the other red objects in the scene. We quantitatively compare the consistency of the masks to human labels (right column) in Table~\ref{tab:segmask_quality}.}
    \label{fig:labels}
\end{figure*}

\subsubsection{Lacquer-based paint} This is a viscous paint consisting of fluorescent powder dissolved in a lacquer which dries clear and glossy. The red and blue is completely dissolved, yielding a glassy finish, while the green is only partially dissolved with some suspended particles. The green paint thus leaves behind a faint white powder when dried~\cite{gloeffex}. This paint can be thinned with standard lacquer thinner, making it more suitable for applying to deformables without stiffening.
\subsubsection{Dye} This substance is a watery staining fluid which works well on absorbent materials. There is one type which is completely transparent under visible light and fluoresces blue, and a variety of other fluorescent colors which have color under visible light. The clear variant is completely invisible under standard lighting, while the colored dyes are only invisible on materials colored similarly. On light colored materials, these dyes can be diluted to further minimize staining and save cost~\cite{darklightfx}. 
\subsubsection{Water-based paint} This paint is acrylic and dries translucent. It is invisible on lighter colored materials but leaves a faint milky residue on darker materials~\cite{gloeffex}.

\subsection{Test Material Properties}
This section describes some important factors to consider when choosing a marking type for a new material. 

\subsubsection{Natural fluorescence} Some materials, particularly white colored papers and cloths, exhibit natural blue fluorescence, prohibiting the use of blue markings on them. Other color fluorescent markings will work on these materials, such as the white towel in our experiment, though the blue fluorescence will shift the color of marks when exposed to UV light. Avoiding materials which naturally fluoresce in the scene is thus desirable for ease of marking and post-processing.
\subsubsection{Fibrous materials} Diluted lacquer paint and dyes are particularly well suited for cloth, whereas undiluted lacquer or water-based paint should be used for solid objects.
\subsubsection{Color} Dark objects result in weaker fluorescence due to light absorption, though in our experience this is typically only significant with near-black materials and cloth.
\subsubsection{Luster} Matte materials are better suited to water-based paint. Shiny objects match the surface finish of lacquer paint.

\subsection{UV Lighting}
We describe the UV lights used to illuminate the setup and the smartplugs which automatically toggle them.

\subsubsection{UV Lights}
\algabbr{} uses 365nm LED UV floodlights to bathe the workspace in light from multiple angles to minimize shadows which fragment segmentation masks. Tube-fluorescent bulbs are \textit{not} usable for this task because they cannot be switched rapidly, while LEDs nearly instantly trigger. Shorter wavelength UV lights yield brighter fluorescence, hence the choice of 365nm light is important over other available 405nm LEDs. \algabbr{} can work in settings with strong ambient room lighting, as in the cable segmentation and towel smoothing datasets, or in controlled lighting settings where room lights can be switched off, as in needle segmentation. Toggleable ambient lights optionally make fluorescent labels stand out more from the background.
\subsubsection{Toggling}
We use an off-the-shelf smart plug which plugs into any standard wall socket, connects to a local network, and is controllable from any device on the same network via a Python interface. This allows programmatically switching lights on and off during data collection, and is a scalable solution for any number of lights by plugging a power adapter into the smart plug.
\subsection{Mask Generation}\label{subsec:mask_generation}
To generate masks, the UV lights are turned on, and if available the ambient white lights turned off. The camera exposure for each sample is found by manually sweeping exposures and selecting the exposure yielding clearest label colors. We use the Zed M stereo camera, which provides a software interface to programmatically control these settings. White-balance is held constant during data collection, and HSV color thresholds are hand-picked from a sample image. For our tasks this calibration process takes only a few minutes, though it could be streamlined by implementing a user interface for automatically picking exposure and thresholds. Figure~\ref{fig:labels} shows examples of extracted masks.

For scenes with both dark and light painted materials, multiple exposures can be captured and post-processed with  HDR~\cite{mertens2007exposure} to retrieve well exposed labels for all colors.
\subsection{Parts List and Cost}
 The total 1-time cost of setting up \algabbr{} for indoor ambient lighting at the time of paper submission is \$282. Based on Amazon's recommended price of \$0.82 per semantic segmentation label on Amazon Mechanical Turk, and using 2 labels per image based on their recommendation for quality~\cite{turk}, this breaks even with Turk at 167 labeled images. In contrast, several of the datasets generated in this work contain well over 1000 labeled images (Section~\ref{sec:experiments}), making the cost of \algabbr{} more than 5x less expensive. 

Parts needed to set up a minimal working system are
\begin{enumerate}
    \item Lights: Everbeam 100W LED 365nm floodlight, available on Amazon for \$88 each.
    \item Smart Plug: Kasa Smart Plug HS103P2, available as a 2-pack on Amazon for \$18.
    \item Camera: Any existing RGB camera will work, however for best results it should have exposure control and manual white-balance options.
    \item Fluorescent Marking: 
    To get started, we recommend beginning with lacquer based paints, being the most versatile and invisible.
    Our paint is sourced from the company ``GloEffex" under the product name ``Transparent UV Paint"~\cite{gloeffex}.
\end{enumerate}

%% file: 4-experiments.tex
\section{Experiments}\label{sec:experiments}


We evaluate \algabbr{} on a set of perception tasks commonly studied in robot manipulation, but \algabbr{} can in principle be applied to other tasks such as keypoint detection. Experiments are designed to evaluate the label quality, data collection rate, and flexibility of \algabbr{} compared to human labeling. Due to the much faster rate of data collection, we are able to increase the difficulty of several tasks compared to prior work by considering more visually challenging scenes including distractor objects. In all experiments, RGB data is collected using a Zed M stereo camera. During training, we use a U-Net~\cite{ronneberger2015u} in the towel corner detection task, and we use a ResNet50-FCN~\cite{long2015fully} for the cable and needle segmentation tasks.

\textbf{Evaluation Metrics:} In the needle and cable segmentation tasks, we compare the labeling quality and throughput of \algabbr{} to human labeling. We report the average intersection over union (\textbf{IOU}) between masks from \algabbr{} and from a human labeler on a set of $10$ training images. We also report the average seconds per label (\textbf{SPL}) for a human labeler and for \algabbr{} to annotate each of these images. We evaluate the quality of the learned network on an unseen test set of images in each domain containing only unpainted objects by reporting the intersection over union (\textbf{IOU}) of the predictions compared to human labels on these test images.

\subsection{Towel Corner Detection for Smoothing and Folding}
Predicting task-relevant keypoints using fully-convolutional neural networks is a popular technique in deformable manipulation applications such as fabric smoothing~\cite{seita_fabrics_2020,ganapathi2021learning}, t-shirt folding~\cite{ganapathi2021learning,kollar2021simnet}, and cable untangling~\cite{grannen2020untangling,viswanath2021disentangling,sundaresan2021untangling}. \citet{seita_fabrics_2020} and \citet{ganapathi2021learning} both predict keypoints corresponding to the corners of a rectangular towel to implement a smoothing policy that iteratively pulls identified corners away from the towel. Both methods are trained on large datasets of simulation data. In this experiment, we collect a dataset of real images containing diverse colors of towels in different configurations with a label on each corner of the towels. We train a neural network to predict towel corners and later use it in an algorithm that smooths and folds towels in the training set.

\subsubsection{Experimental Setup and Assumptions}
The workspace contains an ABB YuMi robot, facing a tabletop with a black tablecloth. We sample towels to place in the workspace from a set of four, monochromatic, rectangular towels. We assume, for this task, that no other objects are in the scene.

\subsubsection{Task Definition}
The goal of this task is to predict masks corresponding to the corners of a towel in the workspace from input images. The predicted masks are used by a heuristic algorithm to smooth and fold the towel. We attempt square double-folds and consider towels ``folded" if the final state of the towel can be pinched at the innermost folded corner and shaken without disrupting the folds in the towel, meaning 4 layers of cloth closely occupy the corner.

\subsubsection{Data Collection}
Self-supervised data collection is a popular technique for training fabric manipulation policies~\cite{hoque2020visuospatial,finn2017deep}. Data for this task is collected autonomously by color thresholding the towel from the black background, picking it up from a random point along the border, shaking it in the air, and dropping it. Because this process often biases towards crumpled states, we manually place the fabric in smoother configurations and collect a small set of images in more orderly configurations as well. Data is collected with the ABB YuMi robot. The training dataset has 3640 images.

\subsubsection{Smoothing and Folding Algorithm} 
\begin{figure}[t]
    \centering
    \includegraphics[width=\columnwidth]{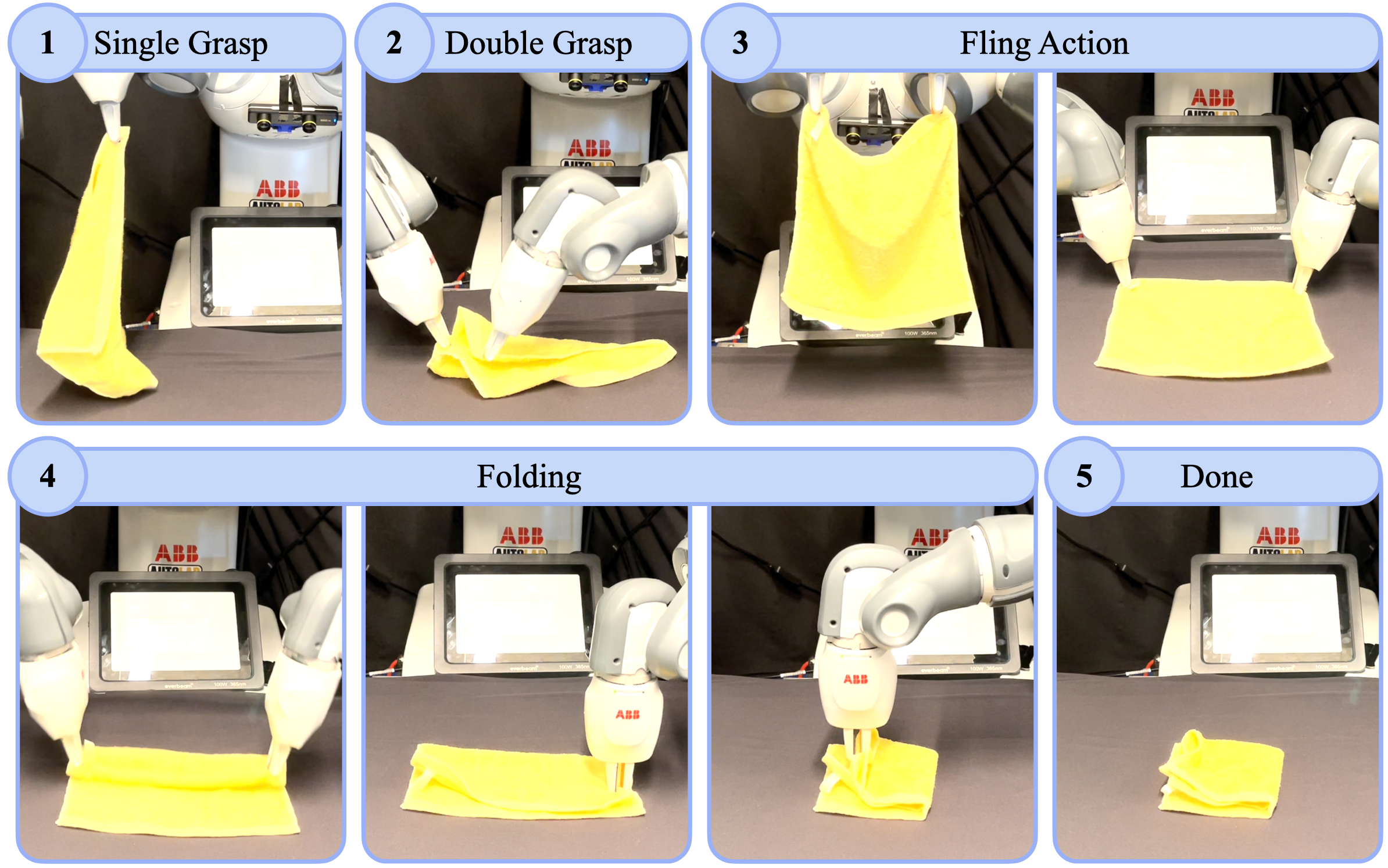}
    \caption{\textbf{Towel Folding:} During smoothing, we use corner predictions from a trained network to implement a heuristic smoothing algorithm. If one corner is visible, the robot drags the towel sideways (\textbf{1}) from this corner to increase others' visibility. If multiple are visible, the robot grabs the two nearest corners (\textbf{2}) and flings (\textbf{3}) across the table. When 4 detected corners are arranged in a square, the robot executes a folding motion using corner positions to compute grasp positions (\textbf{4}), leading to a neatly folded final product (\textbf{5}). Detailed results are shown in table \ref{tab:smoothing_random}.}
    \label{fig:sf}
\end{figure}

Using the corner outputs from the network, we implemented a heuristic algorithm, to first smooth a crumpled towel then fold the smoothed towel, shown in Fig. \ref{fig:sf}. The algorithm repeats \textit{smoothing actions} until it detects the towel is smoothed. During each action, if no corner is detected, the robot randomly resets the cloth by grasping at a random point, shaking and dropping it. If only one corner is detected, the robot grasps the visible corner, shakes it slightly and drags it across the table to spread other corners out. If more than one corner are detected, the robot grasps the pair of corners closest to each other, lifts them up and flings them forwards and back, flattening the towel.

This process terminates when 4 corners are detected whose pairwise distances match the physical size of the towel within a standard deviation of their mean.

After the towel is smoothed, as shown in Fig. \ref{fig:sf} panel 3, the locations of all 4 corners are measured and used to execute an open-loop folding motion. First, the robot grasps the two corners near the robot and puts them on top of the further two corners. Then, a single arm grasps the short folded segment and places it to match the antipodal one, completing the fold. 

\begin{figure}[t]
    \centering
    \includegraphics[width=\columnwidth]{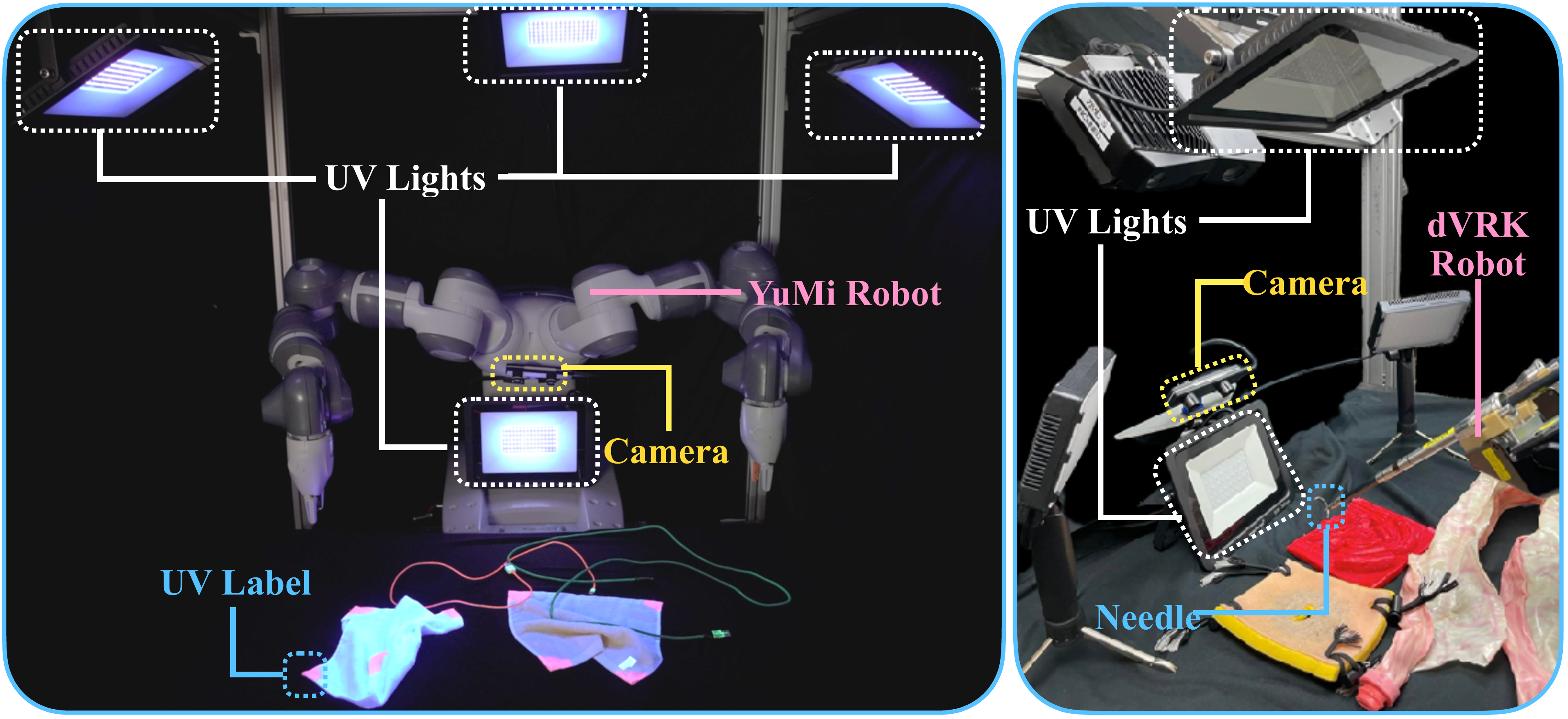}
    \caption{\textbf{Experimental Setups:} We use \algabbr{} on two robot setups. \textbf{Left:} The first, a bimanual YuMi robot, consists of 4 UV lights oriented at different angles to mazimize UV coverage, with a camera mounted between the arms to minimize arm occlusions. All 4 UV lights are toggled with the same smart plug. \textbf{Right:} The second, a dVRK surgical robot, has 2 UV lights and 2 visible LED lamps which are each controlled by separate smartplugs. The UV lights are turned on when the visible lamps are turned off and vice versa.}
    \label{fig:setup}
\end{figure}

We use the depth output from the Zed stereo camera to retrieve the 3D positions of the corners by taking the median of deprojected depth points inside the network output mask, and execute grasps using a fixed gripper orientation.

\begin{figure}[!htb]
    \centering
    \includegraphics[width=\columnwidth]{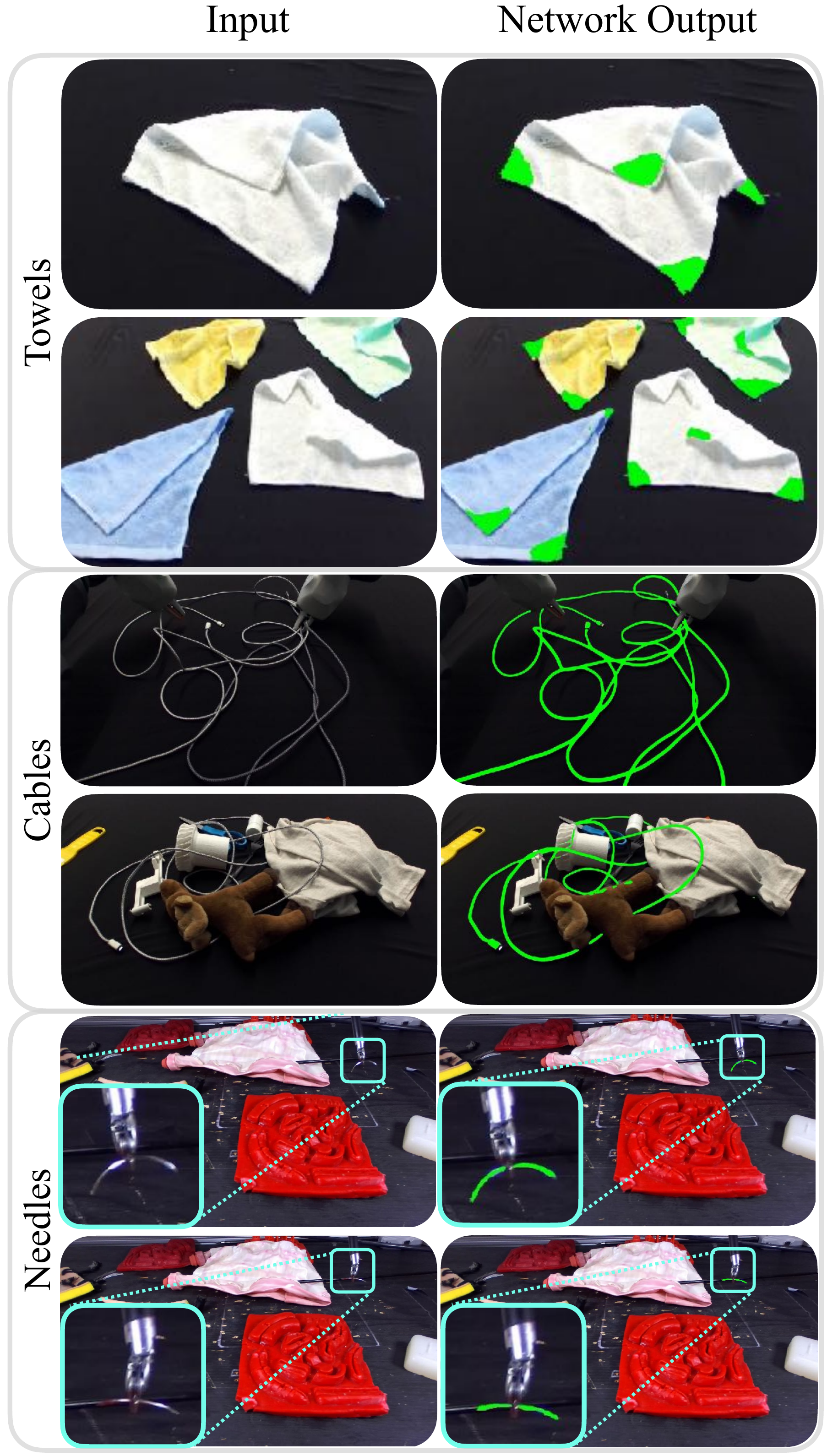}
    \caption{\textbf{Network Predictions on Unseen, Test Images}: We evaluate the trained networks for each of the tasks on images unseen during training. The input images are in the left column, and the right column depicts the input images with the predicted segmentation masks overlaid in green. To test generalization, some of the towel images contain multiple towels, even though all of the training images only had a single towel.  The leftmost cable test image contains two cables, even though all of the training images only contained a single cable.}
    \label{fig:preds}
\end{figure}


\subsubsection{Results}
We evaluate the algorithm on towels in the train set, with random initializations as in autonomous data collection. We execute at most 10 smoothing actions before considering the rollout a failure. Smoothing is successful if it terminates autonomously and proceeds to folding. 

Results are reported in table \ref{tab:smoothing_random}. Smoothing succeeds on average 92\% of rollouts across all towels, and folding on average 83\%. The majority of failure cases are from manipulation challenges and the heuristic algorithm, such as timeouts from repeated grasp failures, or looping because the algorithm grasps diagonal corners over and over, rather than corners detection failure.

\subsection{Cable Segmentation}
Prior work in cable manipulation~\cite{grannen2020untangling,sundaresan2021untangling,viswanath2021disentangling} often assumes that the cable is visually distinguishable from the background via color segmentation. But cables in household or industrial settings may not be chromatically distinct from background objects. This motivates a learning-based approach to predict cable segmentation masks from images. However, labeling cable segmentation masks is extremely tedious due to the complexity of cable configurations.
We apply \algabbr{} to the task of cable segmentation, and include configurations with distractor objects and multiple cables in the scene. Generating these complex segmentation labels via \algabbr{} takes less than 178 ms per image (Table~\ref{tab:segmask_quality}).

\subsubsection{Experimental Setup and Assumptions}
The workspace contains a bilateral ABB Yumi robot. We sample cables to place in the scene from a set of  2 micro-USB to USB cables and a lightning to USB cable. We found that one of the micro-USB cables is naturally fluoresces blue under UV radiation without any painting, so we use this cable in both the training and execution images since we do not paint it. For the other cables, we use a painted version for training images and an unpainted version for execution images. To increase both the difficulty of the task and robustness of the model, we place unpainted distractor objects in the workspace. Several of the items are selected to contain reflections and colors similar to the cables used. At execution time, we introduce several novel distractor objects unseen in training images.

\subsubsection{Task Definition}
The goal of this task is to predict a semantic segmentation mask $I_{\rm seg, cable}$ corresponding to all of the cables in an input image of the workspace.

\subsubsection{Data Collection}
We collect data separately for each of the painted cables. We place each cable in the scene and manually randomize its configuration, knot structure, and position of the YuMi arms. We also randomly place distractor items in the scene.
We collect $486$ labeled images. The bulk of the time was spent manually randomizing the cable, object, and robot configurations.

\subsubsection{Results}
Due to the visual complexity of these scenes, collecting human annotated images was extremely time consuming, with each image taking an average of $446$ seconds to label (Table~\ref{tab:segmask_quality}). \algabbr{} takes an average of $0.178$ seconds to annotate each image, which is \textit{$2511$ times faster}. Labeling all of the image in the dataset with \algabbr{} takes about $87$ seconds \textit{in a single thread}, whereas we estimate from Table~\ref{tab:segmask_quality} that labeling all of the collected images would take over $446 \times 486 = 60$ hours. We find that the mean IOU between \algabbr{} labels and human labels on 10 training images is 0.787 (Table~\ref{tab:segmask_quality}) and the mean IOU between the \algabbr{}-trained segmentation network and human labels on a test set of 10 images is 0.755 (Table~\ref{tab:segmask_quality}).

\subsection{Needle Segmentation}
Segmenting surgical needles in images is a common perception task in surgical robotics research. Some prior works rely on painting the needle and performing color segmentation~\cite{sen2016automating,wilcox2021learning,chiu2020bimanual} or assuming that the rest of the scene is visually distinct from the needle~\cite{sundaresan2019automated}. Recent work uses fully-convolutional neural networks to predict these masks, using a combination of simulation and real data~\cite{wilcox2021learning}, but restrict their task to black backgrounds. In this task, we apply \algabbr{} to needle segmentation in the presence of tissue phantoms and distractor objects in the scene.

\subsubsection{Experimental Setup and Assumptions}
We collect data in a workspace with a bilateral da Vinci Research Kit surgical robot~\citet{dvrk2014} and an inclined Zed M stereo camera. We collect training data using a set of 2 surgical needles and test data using unpainted versions of these needles. In contrast to prior work, which considers a pristine black background for color segmentation~\cite{wilcox2021learning}, we increase the difficulty and realism of the task by randomly placing surgical phantoms, training equipment, and tools in the scene.

\subsubsection{Task Definition}
The goal of this task is to predict a semantic segmentation mask for input images corresponding to all surgical needles.

\subsubsection{Data Collection}
Unlike \cite{wilcox2021learning}, we exclusively train methods on real data, and we use their self-supervised data collection policy to move the needle around the workspace when grasped by the robot's end effector. We periodically insert, remove, and move surgical tissue phantoms in the background like silicone suture practice pads and simulated gut. We use the da Vinci Research Kit for data collection.
The needle dataset consists of 1364 images, with infrequent human interventions to periodically change the poses of the needle in the gripper and the background objects.

\begin{table}[t]
\caption {\textbf{Labeling Technique Comparison:} We evaluate \algabbr{} on a set of 3 robot perception tasks. \textbf{Top half:} We compare the consistency of the UV training labels with human labels as a measure of label quality by comparing their intersection over union (IOU). We report the seconds per label (S.P.L) for both a human labeler and \algabbr{}. On the segmentation tasks, we observe that the training masks for \algabbr{} have an IOU of 0.787 and 0.683 with respect to human labeled masks. Because the cables and needles are very thin, small discrepancies can significantly impact the IOU score negatively, and the labels were also challenging for a human to label. On the needle task, we quantify label quality by using them for pose estimation in Table~\ref{tab:pose_diff}. We observe that \algabbr{} takes 180-2511$\times$ less time than the human to label images. Labeling the entire cable dataset by hand would take approximately 60 hours, whereas it takes \algabbr{} 87 seconds in a single-thread. \textbf{Bottom half:} We evaluate the models trained with \algabbr{} labels on unpainted test images and compare the predictions to human labels. We find that the predictions have an average intersection over union (IOU) of 0.755 and 0.666 on the two tasks. Because cables and needles have thin segmentation masks, small discrepancies with respect to human labels can lead to large negative drops in IOU. We present qualitative analyses of the predictions (Fig.~\ref{fig:preds}).}
\begin{center}
\resizebox{\columnwidth}{!}{%
\begin{tabular}{l|r |rr}
\toprule
& IOU & SPL (human) & SPL (\algabbr{})\\
\hline
\hline
Towel Corner Detection (Train) &  N/A & 22.5 & \textbf{0.125}\\
Cable Segmentation (Train) & 0.787 & 446 & \textbf{0.178} \\
Needle Segmentation (Train) &  0.683 & 40 &  \textbf{0.221}\\
\hline
\hline
Cable Segmentation (Execution) & 0.755 & N/A & N/A\\
Needle Segmentation (Execution) &  0.666 & N/A & N/A\\
\bottomrule
\end{tabular}}
\end{center}
\label{tab:segmask_quality}
\vspace{-4pt}
\end{table}



\begin{table}[t]
\caption {\textbf{Smoothing and Folding Results}. For each towel, 6 trials are conducted with random initial states leading to 24 trials in total. Mean and standard deviation of number of smoothing actions, smoothing and folding success rate are reported.}
\begin{center}
\resizebox{\columnwidth}{!}{%
\begin{tabular}{cccc}
\toprule
Towel & No. Smooth Action  & Smooth Success & Fold Success  \\
\hline
\hline
Blue & $4.4 \pm 1.95$ &	83\% &	67\% \\	
Green & $5.0 \pm 3.16$ & 100\% & 100\% \\	
White & $1.8 \pm 0.41$ &	100\% &	100\% \\	
Yellow & $4.0 \pm 2.91$ &	83\% &	67\% \\	
\hline
\hline
Average & $3.7\pm 2.52$ &	92\% &	83\% \\	
\bottomrule
\end{tabular}}
\end{center}
\label{tab:smoothing_random}
\end{table}






\begin{table}[t]
\caption {\textbf{\algabbr{} Needle Pose Estimate Consistency. Top Row:} We use the training needle masks (Training Labels) generated by the UV labels to estimate the needle's pose using the 3D needle pose reconstruction algorithm from \citet{wilcox2021learning}. We compare the pose to the pose generated by human labels of the training images. We find that the pose estimate is within 1.7mm and 6.9$^\circ$ the poses generated by human labels on average. In this test, we ensure that the images do not contain the needle in pathological cases, as described in \citet{wilcox2021learning}. \textbf{Bottom Row:} We evaluate the trained network predictions on test images in the same way, and find that the resulting average pose estimate is within 1.7mm and 8.8$^\circ$ of the pose generated by the human labeled segmasks.
}
\begin{center}
\resizebox{0.85\columnwidth}{!}{%
\begin{tabular}{l|r|r}
\toprule
& Position Difference(mm) & Rotation Difference\\
\hline
\hline
Training Labels & $1.7$mm & $6.9^\circ$\\

Test Predictions & $1.7$mm & $8.8^\circ$\\
\bottomrule
\end{tabular}}
\end{center}
\label{tab:pose_diff}
\end{table}

\subsubsection{Results}
We compare the annotations generated by \algabbr{} to human annotations on a set of 38 training images. The mean IOU between the segmentation masks in the two sets is 0.683. We train a segmentation network whose prediction has a mean IOU of 0.666 with respect to human annotations on an unpainted test set of 20 images. While this seems relatively low, this is due to the very thin profile of needles, and slight variations in the human and \algabbr{} annotations can significantly affect IOU. The masks are qualitatively very similar as shown in Fig. \ref{fig:labels}, and we quantify this by running the needle pose reconstruction algorithm from \citet{wilcox2021learning} on stereo images in both the training and test set. We limit test images to those where both tips of the needle are visible. The resulting average pose error between labels from \algabbr{} and humans is 1.7mm and $6.9^\circ$, and the pose error on the test set between network outputs and human labels is 1.7mm and $8.8^\circ$.

%% file: 5-discussion.tex
\section{Discussion}
We present \algname{}, a framework for collecting segmentation labels and keypoints without human labeling. \algabbr{} can be applied to augment existing self-supervised dataset collection techniques in robot manipulation domains. In future work, we hope to investigate other applications of \algabbr{}, such as using fluorescent markers to obtain object poses, detecting garment edges, or using UV markers for online dense reward assignment in RL tasks. In addition, LUV could be extended to produce a larger  diverse dataset of fabric annotations on RGB images.

%% file: 6-acknowledgement.tex
\section*{Acknowledgments}
\textcolor{black}{\footnotesize{This research was performed at the AUTOLAB at UC Berkeley in affiliation with the Berkeley AI Research (BAIR) Lab, the CITRIS "People and Robots" (CPAR) Initiative, the Real-Time Intelligent Secure Execution (RISE) Lab and UC Berkeley's Center for Automation and Learning for Medical Robotics (Cal-MR). This work is supported in part by donations from Intuitive Surgical, Google, Siemens, Autodesk, Bosch, Toyota Research Institute, Honda, Intel, Hewlett-Packard and by equipment grants from PhotoNeo and NVidia. We thank Adam Lau for his photography and Kishore Srinivas for assistance with data labelling.}}